# Rüzgar Hızının Yapay Sinir Ağları ve ANFIS Metotları Kullanılarak Tahmin Edilmesi (Ölçüm Şamandırası Örneği)

# Prediction of Wind Speed Using Artificial Neural Networks and ANFIS Methods (Observation Buoy Example)

Timur İnan[1], Ahmet Fevzi Baba[2]
[1]Elektrik ve Enerji Programı, İstanbul Arel Üniversitesi, İstanbul, Türkiye
timurinan@arel.edu.tr
[2]Elektrik - Elektronik Mühendisliği Bölümü, Marmara Üniversitesi, İstanbul, Türkiye
fbaba@marmara.edu.tr

*Özetçe*—Rüzgâr hızının tahmin edilmesi; gemilerin rota belirlemesi, rüzgar güllerin verimli kullanılması, tarım faaliyetlerinin doğru planlanması gibi bir çok konuda önemli rol oynamaktadır. Bu çalışmada yapay sinir ağları (YSA) ve uyarlanabilir yapay sinir bulanık sistemi (ANFIS) yöntemleri kullanılarak rüzgar hızı tahmini hesaplaması yapılmıştır. Tahmin yapılabilmesi için gerekli olan veriler POSEIDON şamandıra sisteminin içindeki bir şamandıra olan E1M3A isimli şamandıradan temin edilmiştir. Ortaya konulan YSA, 3 katmanlı, 50 nörona, 6 giriş ve 1 çıkışa sahip, harici girişli, doğrusal olmayan otomatik regresif (Nonlinear Auto Regressive with External Input (NARX)) tipte bir yapay sinir ağıdır.Ortaya konulan ANFIS sistemi ise 6 girişli, 1 çıkışlı, giriş başına 3 adet üyelik fonksiyonuna (Membership Function (MF)) sahip bir bulanık çıkarım sistemidir. Ortaya konulan sistemler, 3 saat sonrası rüzgar hızı tahmini yapabilmeleri için eğitilerek tahmin verileri elde edilmiş ve elde edilen değerler gerçek ölçümlerle karşılaştırılarak sistemlerin başarıları ortaya konulmuştur. Sistemlerden elde edilen tahmin değerlerinin başarılarının değerlendirilmesinde hataların karelerinin ortalaması (Mean Squared Error(MSE)) ve tahminler ile beklenen değerler arasındaki benzerlik (Regression (R)) değerleri kullanılmıştır. Tahmin sonuçlarına göre YSA, eğitimde 2.19 MSE ve 0.897 R değerlerini, doğrulamada 2.88 MSE ve 0.866 R değerlerini, testte ise 2.93 MSE ve 0.857 R değerlerini elde etmiştir. ANFIS yöntemi ise 0.31634 MSE ve 0.99 R değerlerini elde etmiştir.

*Anahtar Kelimeler: Rüzgâr Hızı Tahmini; Yapay Sinir Ağı; ANFIS*

*Abstract*—Estimation of the wind speed plays an important role in many issues such as route determination of ships, efficient use of wind roses, and correct planning of agricultural activities. In this study, wind velocity estimation is calculated using artificial neural networks (ANN) and adaptive artificial neural fuzzy inference system (ANFIS) methods. The data required for estimation was obtained from the float named E1M3A, which is a float inside the POSEIDON float system. The proposed ANN is a Nonlinear Auto Regressive with External Input (NARX) type of artificial neural network with 3 layers, 50 neurons, 6 inputs and 1 output. The ANFIS system introduced is a fuzzy inference system with 6 inputs, 1 output, and 3 membership functions (MF) per input. The proposed systems were trained to make wind speed estimates after 3 hours and the data obtained were obtained and the successes of the systems were revealed by comparing the obtained values with real measurements. Mean Squared Error (MSE) and the regression between the predictions and expected values (R) were used to evaluate the success of the estimation values obtained from the systems. According to estimation results, ANN achieved 2.19 MSE and 0.897 R values in training, 2.88 MSE and 0.866 R values in validation, and 2.93 MSE and 0.857 R values in testing. ANFIS method has obtained 0.31634 MSE and 0.99 R values

## I. GİRİŞ

Rüzgâr hızının tahmini konusu, rüzgâr hızının etken faktör olduğu alanlarda oldukça önem taşımaktadır. Bu alanlardan bazıları, rüzgârdan elektrik üretimi, gemi rota planlaması ve tarım alanlarıdır.

Bu çalışmada rüzgâr hızı tahmini yapabilme amacıyla iki ayrı yöntem kullanılmış ve oluşturulan sistemlerin başarıları karşılaştırılmıştır.

Çalışmada kullanılan bilgiler IFREMER sisteminden elde edilmiştir [1]. IFREMER sistemi, dünya üzerindeki birçok ölçüm şamandırasından alınan bilgileri internet sitesi üzerinden paylaşmaktadır. Çalışmada kullanılan bilgiler ise POSEIDON sistemine ait E1M3A şamandırasından elde edilmiştir [2]. İşlenen veriler, E1M3A şamandırasından alınan 2011-2018 yılları arasına ait verilerdir.

Tahmin yapabilmek amacıyla sıcaklık, basınç, rüzgar hızı, basınç farkı, rüzgar hızı değişimi, sıcaklık değişimi gibi veriler girdi olarak alınmıştır. Şamandıradan alınan veriler 3 saatlik aralıklarla sağlandığından ortaya konulan tahmin sistemleri de 3 saat sonrası için tahmin yapmaktadır.

Tahmin yapılması için kullanılan yöntemler YSA ve ANFIS yöntemleridir.

YSA, günümüzde birçok çalışmada tahmin yöntemi olarak kullanılmaktadır. Özsoy ve Fırat tarafından yapılan bir çalışmada kirişsiz döşemeli betonarme bir binada çeşitli parametrelere bağlı olarak meydana gelebilecek ötelenme miktarının tahmini YSA kullanılarak yapılmıştır [3].

Mazanoğlu, Uşak ili ve çevresindeki depremlerin modellenmesi amacıyla YSA kullanmıştır [4].

Öge, yapay sinir ağları ve regresyon analiz yöntemlerini kullanarak kaya kütle sınıflandırması yapan bir sistem ortaya koymuştur [5].

Toçoğlu ve diğerleri, Türkçe metinlerden duygu analizi çıkarımı yapabilen bir sistem ortaya koymuştur. Sistemde birçok yöntem ile birlikte YSA da kullanılmıştır [6].

ANFIS yöntemi ise yine yapay sinir ağları gibi tahmin amacıyla kullanılan oldukça popüler bir yöntemdir.

Minaz, ortaya koyduğu rüzgâr hızı tahmin sisteminde ANFIS yöntemini kullanmıştır [7].

İnan, ortaya koyduğu robot kolu sisteminde, kameradan tespit edilen hedefin takibini yaptırmış ve hedefin bir sonraki adımda nerede olabileceğini YSA ve ANFIS yöntemleriyle hesaplayarak yöntemlerin başarısını ortaya koymuştur [8].

Rüzgar hızı tahmini için günümüzde birçok tahmin yöntemi kullanılmaktadır. Bu yöntemlerin başında YSA, ANFIS, Destek Vektör Makinaları (SVM), istatiksel yöntemler gibi yöntemler gelmektedir.

Castellanos ve James, saatlik ortalama rüzgâr hızını tahmin edebilen, ANFIS tabanlı bir sistem sunmuşlardır [9].

Mohandes ve diğerleri tarafından ortaya konulan rüzgâr hızı tahmin sisteminde SVM yöntemi kullanılmıştır [10], [11].

Pousinho ve diğerleri, parçacık sürü algoritması ve ANFIS yöntemlerinin birlikte kullanıldığı melez bir sistem ortaya koyarak Portekiz için kısa süreli rüzgâr gücü tahmini yapabilen bir sistem ortaya koymuşlardır [12].

Ramasamy ve diğerleri, YSA tabanlı bir rüzgar tahmin sistemi ortaya koymuşlardır. Ortaya koydukları sistemde Hindistan'ın dağlık bölgelerindeki rüzgar güllerinden elde edilen veriler kullanılmıştır [13].

Dokur ve diğerleri sundukları bildiride, görgül kip ayrışımı ve Elman-Jordan yapay sinir ağı tabanlı hibrit bir rüzgar hızı kestirim sistemi ortaya koymuşlardır [14].

Zeng ve diğerleri, iki aşamalı ANFIS tabanlı bir rüzgar hızı kestirim sistemi ortaya koymuşlardır [15].

Kırbaş, istatistiksel yöntemler ve YSA kullanarak kısa dönem çok adımlı bir rüzgar hızı tahmin sistemi ortaya koymuştur [16].

Inan ve Baba, ortaya koydukları sistemde gemiler için rüzgar yön ve hızı, akıntı yön ve hızı, dalga yön ve yüksekliklerini YSA ile tahmin edebilen bir sistem ortaya koymuşlardır [17].

## II. MATERYAL VE YÖNTEM

Bu çalışmada rüzgar hızı tahmini için YSA ve ANFIS yöntemleri kullanılmıştır. Her iki yöntem için de girişler; sıcaklık (celcius), basınç (milibar), rüzgar hızı (m/s), basınç değişimi (milibar), rüzgar hızı değişimi (m/s) ve sıcaklık değişimidir (celcius). Çıkış ise 3 saat sonrasına ait rüzgâr hızı tahmin değeridir (m/s).

### A. Oluşturan YSA modeli

YSA modelinin oluşturulması için MATLAB istatistik ve optimizasyon araç kutusu kullanılmıştır. Ortaya konulan yapay sinir ağı NARX tipi bir yapay sinir ağıdır.

Şekil 1'de, oluşturulan yapay sinir ağının giriş ve çıkışları görülmektedir.

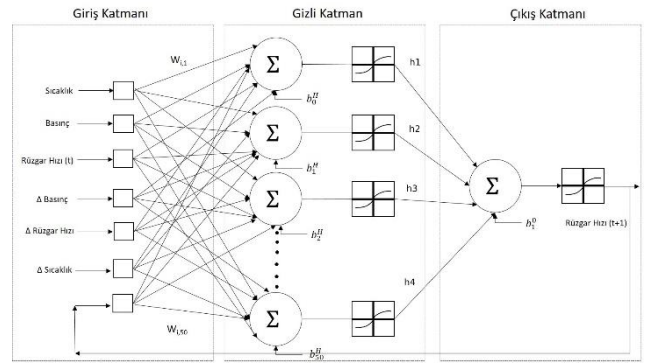

**Şekil 1.** Ortaya konulan yapay sinir ağının yapısı

NARX tipi yapay sinir ağlarının giriş katmanı, gizli katman ve çıkış katmanları arasındaki bağıntılar Denklem 1 ve Denklem 2'de detaylı olarak verilmektedir [18].

$$h_i = \prod \left[ \sum_{j=1}^{\Phi} w_{i,j} x_j + b_i^H \right] \qquad (1)$$

$$y_k = \prod \left[ \sum_{i=1}^{H} w_{K,i} h_i + b_i^o \right] \qquad (2)$$

Burada, hi, i numaralı nöronun çıkış değeri, wi,j, i numarlı nöronun j adımındaki ağırlık değeri, x_j, j adımındaki giriş değeri, b_i^H, i numaralı nöronun çıkışa etki katsayısı, Φ, girişler ile gizli katman arasındaki bağlantıların sayısı, H, nöron sayısı, O yapay sinir ağının çıkış sayısı, y_k, yapay sinir ağının çıkışıdır.

### B. Oluşturulan ANFIS Modeli

Adaptive artificial neural fuzzy inference system (ANFIS) yöntemi ilk kez 1993 yılında Jang tarafından ortaya konulan ve içerisinde hem yapay sinir ağları hem de bulanık mantık yöntemini barındıran bir çıkarım yöntemidir [19]. Oluşturulan ANFIS sistemi Sugeno tipi, 6 girişli ve 1 çıkışlı bir çıkarım sistemidir. Sisteme ait girişler ve çıkış Şekil 2'de görülmektedir.

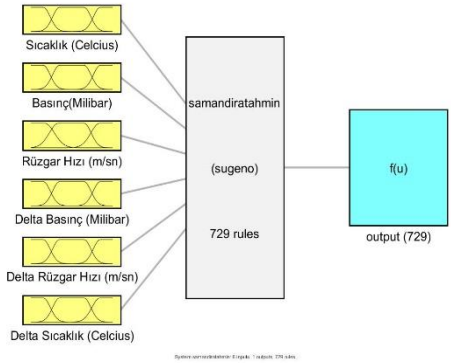

**Şekil 2.** Ortaya konulan ANFIS sisteminin yapısı

6 girişin her biri için üçer adet komşuluk fonksiyonu (Membership Function-MF) tanımlanmıştır. Kullanılan komşuluk fonksiyonu, her giriş için gauss fonksiyonu olarak seçilmiştir.

Şekil 3'te her bir giriş için kullanılan komşuluk fonksiyonlarının detayları verilmiştir.

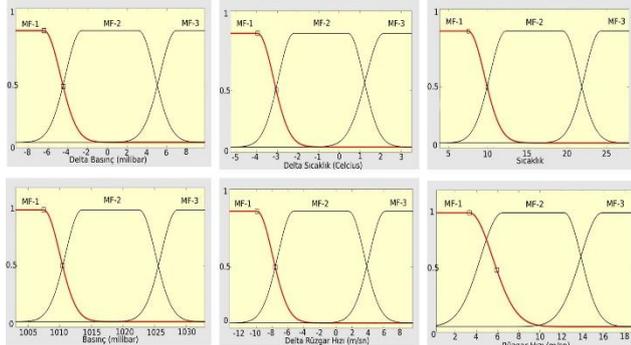

**Şekil 3.** ANFIS sisteminde kullanılan komşuluk fonksiyonları

### III. BULGULAR

Rüzgar hızı tahmini için geliştirilen iki sistem de MATLAB ortamında test edilmiştir. Testlerin gerçekleştirildiği bilgisayar Intel Pentium i7 işlemciye sahiptir. İşlemci 3.6 GHz hızında olup, bilgisayar 8 GB RAM'a sahiptir.

Yapay sinir ağının performansı MSE (Mean Squarred Error) yöntemiyle hesaplanmıştır. Hataların karelerinin ortalaması anlamına gelen bu yöntemde yapay sinir ağının performansını bulmak için kullanılan fonksiyon Denklem 3'te görülmektedir.

$$MSE = \frac{\sum_{t=1}^{n}(e_t - o_t)^2}{n} \qquad (31)$$

$e_t$, $t$ anında eğitim için kullanılan verinin değeri, $o_t$, yapay sinir ağının $t$ anında yaptığı tahmini temsil etmektedir. Eğitim verisi ile tahmin arasındaki fark hatayı vermektedir. $n$ adet veri için teker teker hataların karesi alınmakta ve hataların kareleri toplamı veri miktarına bölünmektedir.

Yapay sinir ağının çıkışının gerçek değerler ile ne kadar benzediğinin tespiti için tahmin ve gerçek değerlerin regresyon değerleri hesaplanmıştır. Regresyon değerlerinin hesaplanmasında Denklem 4 kullanılmıştır.

$$R = \frac{\sum_{t=1}^{n}(e_t - \bar{e}_t)(o_t - \bar{o}_t)}{\sqrt{\sum_{t=1}^{n}(e_t - \bar{e}_t)}\sqrt{\sum_{t=1}^{n}(o_t - \bar{o}_t)}} \qquad (4)$$

$e_t$, $t$ anında eğitim verisinin aldığı değeri, $o_t$, $t$ anında yapay sinir ağının yaptığı tahmini, $\bar{e}_t$, eğitim verisinin türevini, $\bar{o}_t$, tahmin verisinin türevini temsil etmektedir.

Oluşturulan yapay sinir ağı, eğitimde 2.19 MSE ve 0.897 R değerlerini, doğrulamada 2.88 MSE ve 0.866 R değerlerini, testte ise 2.93 MSE ve 0.857 R değerlerini elde etmiştir.

Şekil 4'te oluşturulan yapay sinir ağının regresyon değerleri görülmektedir.

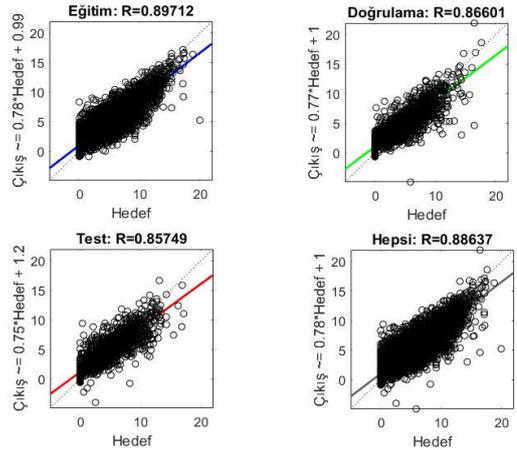

**Şekil 4.** Ortaya konulan yapay sinir ağının regresyon değerleri

Rüzgar hızı tahmini için oluşturulan anfis sisteminin başarısının ölçülmesi ve yapay sinir ağı ile eğitilen sisteme nazaran başarısının kıyaslanabilmesi için yine 3 ve 4 numaralı eşitlikler kullanılmıştır. ANFIS yöntemi ise 0.31634 MSE ve 0.99 R değerlerini elde etmiştir. Şekil 5'te ortaya konulan ANFIS sisteminin başarısı görülmektedir.

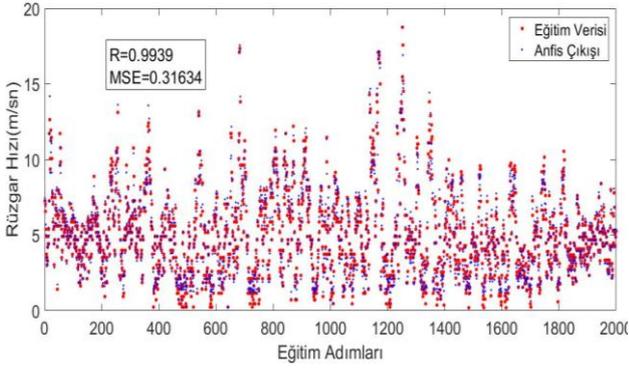

**Şekil 5.** Ortaya konulan ANFIS sisteminin başarısı

Şekil 5'te kırmızı noktalar eğitim verilerini, mavi noktalar ise ortaya konulan ANFIS sisteminin tahmin verilerini ortaya koymaktadır.

## IV. SONUÇLAR

Yapay sinir ağları ve ANFIS yöntemleri birbirleri ile karşılaştırıldığında ANFIS yönteminin yapay sinir ağı yöntemine nazaran daha başarılı tahminler yapabildiği gözlemlenmiştir.

Tahmin sonuçlarına göre YSA, eğitimde 2.19 MSE ve 0.897 R değerlerini, doğrulamada 2.88 MSE ve 0.866 R değerlerini, testte ise 2.93 MSE ve 0.857 R değerlerini elde etmiştir. ANFIS yöntemi ise 0.31634 MSE ve 0.99 R değerlerini elde etmiştir.

Şekil 6'da ortaya konulan sistemlerin başarıları grafiksel olarak gösterilmektedir.

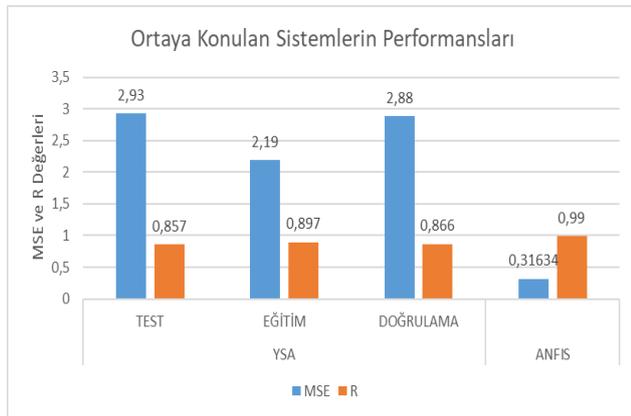

**Şekil 6.** Ortaya konulan sistemlerin performansları

Şekil 6'dan da anlaşılacağı üzere ANFIS yöntemi, yapay sinir ağları yöntemine göre daha az hatalı tahminler yapabilmiş ve yaptığı tahminler ise gerçek verilere oldukça yakın çıkmıştır. Bu çalışmada, deniz ölçüm şamandırasından alınan bilgiler kullanılarak rüzgar hızı tahmini yaptırılmıştır. Tahmin için YSA ve ANFIS yöntemleri kullanılmıştır. Ortaya konulan iki sistemin başarısı karşılaştırılmış ve ANFIS yöntemi daha başarılı bulunmuştur.